\documentclass[11pt,a4paper]{article}
\usepackage{times,latexsym}
\usepackage{url}
\usepackage[T1]{fontenc}
\usepackage{amsmath}
\usepackage{booktabs}
\usepackage{graphicx}
\usepackage[inline]{enumitem}
\usepackage[acceptedWithA]{tacl2021v1}
\hypersetup{hidelinks,
  pdftitle={Matching the Training-Time Weight Rule in Inference-Time Expert Deactivation},
  pdfauthor={Hao Cheng, Hanzhang Jia, Zhiyu Yang, Bo Ma}}
\usepackage[font=small,skip=3pt]{caption}

\title{Matching the Training-Time Weight Rule in Inference-Time Expert Deactivation}

\author{\parbox{0.92\textwidth}{\centering
Hao Cheng\textsuperscript{1}\quad Hanzhang Jia\textsuperscript{2,*}\quad Zhiyu Yang\textsuperscript{3}\quad Bo Ma\textsuperscript{1,*}\\[2pt]
\textsuperscript{1}Zhejiang Gongshang University\quad \textsuperscript{2}University of Sussex\quad \textsuperscript{3}Chang'an University\\[2pt]
\texttt{hj303@sussex.ac.uk}\quad\texttt{mabo@mail.zjgsu.edu.cn}\\[2pt]
{\small \textsuperscript{*}Corresponding authors: Hanzhang Jia (lead corresponding author) and Bo Ma.}}}
\date{}

\AtBeginDocument{\lefthyphenmin=2\righthyphenmin=3}
\begin{document}
\maketitle

\begin{abstract}

The performance impact of expert deactivation depends on both the disabled set and the routing and weighting of the remaining experts. We compare removal, freezing, remaining-expert renormalization, and unnormalized re-routing on identical disabled sets across five instruction-tuned checkpoints from four model families, with a base OLMoE checkpoint as an additional comparison. A weight-sum model separates the loss of disabled contributions from changes in expert-branch scale and predicts the highest-loss implementation from the training-time weight rule, with remaining-expert renormalization for raw-probability models and unnormalized re-routing for normalized-weight models. All five checkpoints match this prediction. Matched survivor sets isolate weight handling. On one DeepSeek-V2-Lite disabled set, freezing yields a perplexity of 17.21 while renormalization yields 258,621.64. At 50 percent deactivation using activation-based selection, perplexity reaches 1.4 to 73 times the intact baseline under the training rule and 14 to 28,740 times the baseline under a rule mismatch. The same model extends to inference-time width, where the penalty from additional renormalization grows as the number of active experts decreases, and OLMoE exhibits this trend. Multiple-choice and free-form evaluations separate likelihood collapse from answer accuracy and reveal differences in generation behavior. Training-time weight rules are therefore a practical reference for expert pruning, offloading, and width adjustment.

\end{abstract}

\section{Introduction}

Two deactivation runs can preserve exactly the same surviving experts and nevertheless produce perplexities separated by four orders of magnitude. On the same DeepSeek-V2-Lite disabled set, freezing gives a perplexity of 17.21, compared with 258,621.64 under remaining-expert renormalization. The difference lies in a single choice, namely how the surviving expert weights are treated.

Open-weight mixture-of-experts (MoE) models distribute their parameters across expert layers while activating only a small subset of experts for each token. OLMoE, DeepSeek-V2, Granite, and Qwen3 follow this architecture \citep{muennighoff2025olmoe,deepseek2024v2,granite2024,qwen2025}. Sparse activation reduces computation per token, but the full expert bank still incurs storage, loading, and scheduling costs. Expert pruning, offloading, and dynamic routing address these deployment costs by changing which experts are available or selected \citep{lu2024eepa,lasby2025reap,chen2026dynamic}. Deactivating experts requires explicit choices about the existing token assignments and their associated weights. Removal masks the disabled experts, re-routes their tokens, and recomputes weights using the model's original rule. The resulting change therefore combines expert substitution with weight redistribution. Freezing preserves the original routing decisions and surviving weights, removing only the disabled experts' weighted contributions. Affected tokens consequently retain fewer effective experts and a smaller total weight. Remaining-expert renormalization also leaves the vacated slots unfilled, but rescales the surviving weights to sum to one, combining contribution loss with additional scaling. Unnormalized re-routing fills the vacated slots and weights the replacement experts by their raw router probabilities. It preserves the number of active experts while reducing expert-branch gain in models trained with normalized weights. The disabled set determines which contributions disappear, and the implementation determines how the remaining computation is weighted.

Existing studies primarily compare expert-selection criteria within a fixed deactivation implementation \citep{lu2024eepa,lasby2025reap,chen2026dynamic}. \citet{specialization2024} compare removal and freezing without distinguishing models by their training-time weight rule. \citet{chen2026halving} examine the normalization reference when reducing the number of active experts, but do not compare all four implementations on an identical disabled set. The performance differences attributable to implementation, and the correspondence between the most damaging implementation and the training rule, therefore remain unresolved. This paper makes the deactivation implementation itself a first-class experimental variable and links its effect to the weight rule used during training.

We develop a weight-sum model covering all four implementations, using the training-time computation as the reference for analyzing contribution loss and expert-branch scaling. Removal recomputes weights under the original rule, whereas freezing retains the surviving experts' original weights. Remaining-expert renormalization and unnormalized re-routing instead prescribe normalized weights and raw probabilities, respectively, so their compatibility with training depends on the model. In OLMoE and DeepSeek-V2, the selected experts' raw probabilities already sum to less than one during training. Renormalizing the survivors raises the retained sum to one and amplifies their contribution relative to the residual branch. In Granite and Qwen3, the selected weights sum to one during training. Removing this normalization reduces the weight sum below one and uniformly attenuates the weighted output for a given selected set. The model thus predicts that, for a fixed model and disabled set, remaining-expert renormalization will produce the largest perplexity increase in raw-probability models, and unnormalized re-routing will produce the largest increase in normalized-weight models.

Our experiments cover five instruction-tuned checkpoints from four families, with a base OLMoE checkpoint as an additional comparison. For each model and deactivation proportion, all four implementations use the same disabled set. We then vary the selection criterion to examine the effect of the set itself. Model parameters remain fixed throughout, and evaluation includes WikiText-2 perplexity, multiple-choice tasks, and free-form generation on GSM8K. When half of the experts are deactivated using activation-based selection, the highest perplexity occurs under remaining-expert renormalization for OLMoE and DeepSeek-V2-Lite, and under unnormalized re-routing for both Granite checkpoints and Qwen3, as predicted. Implementations that retain the training rule yield perplexities between 1.4 and 73 times the intact baseline; mismatched implementations yield between 14 and 28,740 times the baseline. The disabled set changes the magnitude of the loss within a fixed implementation, as the Granite grids in the appendices show. On instruction-tuned OLMoE at 50\% deactivation, implementations that retain the training rule remain above chance on some two-option tasks, while MMLU approaches chance and GSM8K strict-match accuracy is zero under every implementation evaluated. The weight-sum model also extends to changes in inference-time width. For fixed router probabilities, selecting fewer experts lowers their total probability and increases the amplification introduced by renormalization. The model consequently predicts a larger penalty from a weight-rule mismatch at smaller widths. OLMoE exhibits this trend, with the perplexity penalty from renormalization growing as the number of active experts decreases.

Our contributions are threefold:

\begin{enumerate}[leftmargin=1.4em,topsep=2pt,itemsep=1pt]

\item We establish the performance differences among four deactivation implementations on identical disabled sets and identify the high-loss implementation associated with each training-time weight rule.

\item We develop a weight-sum model of all four implementations that explains the gain changes introduced by adding or removing normalization and predicts which implementation produces the most severe degradation in each model class.

\item We extend the model's predictions to changes in the number of active experts and verify the predicted trend on OLMoE, where additional renormalization causes increasingly severe perplexity degradation at smaller widths. This result extends the explanation based on weight-rule mismatch from fixed expert deactivation to inference-time width adjustment.

\end{enumerate}

\section{Background and Deactivation Implementations}
\label{sec:background}

Given a token representation $h$, the router assigns a score $z_i$ to each of $N$ experts and applies a softmax over the full expert bank:

\begin{equation}
p_i = \frac{\exp(z_i)}{\sum_{j=1}^{N} \exp(z_j)}.
\end{equation}

The $k$ experts with the highest probabilities form the active set $T$. Let $y_i$ denote the output of expert $i$ and $g_i$ its weight. The layer output combines the residual with the weighted expert outputs:

\begin{equation}
\label{eq:layer}
o = h + \sum_{i \in T} g_i\, y_i .
\end{equation}

The training-time weight rule determines how $g_i$ is computed. OLMoE and DeepSeek-V2 use the raw probabilities, so the selected experts' weights sum to $m_T$, which is less than one \citep{muennighoff2025olmoe,deepseek2024v2}:

\begin{equation}
g_i = p_i, \qquad m_T = \sum_{i \in T} p_i < 1.
\end{equation}

Granite and Qwen3 normalize the weights within the selected set \citep{granite2024,qwen2025}. Writing $s_T$ for the sum of that set's raw probabilities gives:

\begin{equation}
g_i = \frac{p_i}{s_T}, \qquad s_T = \sum_{j \in T} p_j, \qquad m_T = 1.
\end{equation}

\begin{figure*}[t]
\centering
\includegraphics[width=0.85\textwidth]{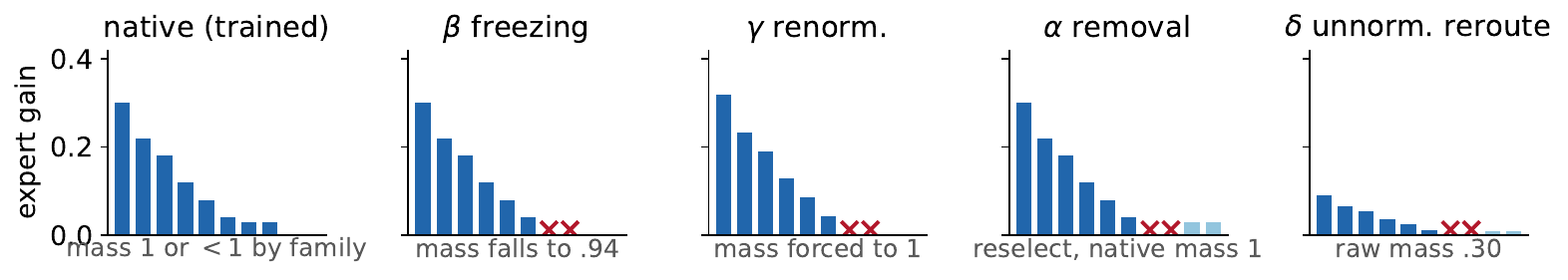}
\caption{Four implementations in a ten-expert example. The original top-8 set occupies slots 0-7, with slots 6 and 7 deactivated (red crosses). Re-routing fills the vacated slots with replacement experts, freezing retains the surviving weights, and renormalization rescales them. Unnormalized re-routing uses the original router probabilities.}
\label{fig:overview}
\end{figure*}

Both rules retain the residual term $h$; $m_T$ describes the total weight on the expert branch. Let $D$ be the disabled set. Assigning zero effective weight to unselected experts, $m_D$ is the total weight of the originally active experts in $D$. Table~\ref{tab:implementations} summarizes the four implementations, and Figure~\ref{fig:overview} illustrates their routing and weighting operations.

\begin{table}[t]
\centering
\small
\setlength{\tabcolsep}{2pt}
\begin{tabular}{@{}lll@{}}
\toprule
Implementation & Weight rule & Weight sum \\
\midrule
$\alpha$ removal & re-select, training rule & $\sum_{i \in T'} q_i$, 1 \\
$\beta$ freezing & keep trained weights & $m_T - m_D$ \\
$\gamma$ renorm. & rescale & 1 \\
$\delta$ unnorm. & re-select, $p_i$ & $s_{T'}$ \\
\bottomrule
\end{tabular}
\caption{Routing, weight computation, and weight sums for the four implementations. $D$ is the disabled set, $m_D$ its originally active weight, and $T'$ the re-selected set. The probabilities $q_i$ are recomputed over the remaining bank; $s_{T'}$ sums the original probabilities of the re-selected experts. The two sums for $\alpha$ correspond to raw-probability and normalized-weight models, respectively.}
\label{tab:implementations}
\end{table}

Removal with re-routing ($\alpha$) masks disabled experts before the softmax and selects a new top-$k$ set $T'$ from the remaining bank. Raw-probability models use the recomputed probabilities $q_i$, whereas normalized-weight models continue to normalize within $T'$. Removal therefore follows the original weight-computation rule while changing both the selected experts and their weights. This is a common implementation in expert pruning \citep{lu2024eepa,lasby2025reap}.

Zero-contribution freezing ($\beta$) preserves the original routing decisions and sets the disabled experts' contributions to zero. Surviving experts retain their original weights. Remaining-expert renormalization ($\gamma$) selects the same survivors, but scales their weights proportionally to sum to one. Neither implementation introduces replacement experts; their difference is whether the surviving contribution is rescaled.

Unnormalized re-routing ($\delta$) excludes disabled experts at selection time, fills the active set with replacements, and weights the selected experts by their original softmax probabilities $p_i$. Disabled experts still contribute to the softmax denominator. This distinguishes $\delta$ from $\alpha$, which masks experts before the softmax is computed.

\subsection{A Weight-Sum Model of Expert Deactivation}
\label{sec:model}

Removal and freezing follow both training-time weight rules. Remaining-expert renormalization is the native rule of normalized-weight models, and unnormalized re-routing is the native rule of raw-probability models. Applying either outside its native family creates a weight-rule mismatch.

Taking the intact layer in Equation~\eqref{eq:layer} as the reference, we analyze how deleting expert contributions and rescaling their weights affect the layer output. For removal, let $p_D$ denote the disabled experts' total probability over the full bank. Masking them gives:

\begin{equation}
q_i = \frac{p_i}{1-p_D}, \qquad i \notin D,
\end{equation}

The router then selects $T'$ using $q_i$. The resulting output depends on both the outputs of the newly selected experts and their weights. The weight sum describes the scale of this weighted combination, while substitution changes the expert outputs that enter it.

Freezing retains the originally selected survivors, giving the layer output:

\begin{equation}
o_{\beta} = h + \sum_{i \in T \setminus D} g_i\, y_i .
\end{equation}

Its difference from the intact output is exactly the removed weighted contribution:

\begin{equation}
\Delta_{\beta} = o - o_{\beta} = \sum_{i \in D} g_i\, y_i .
\end{equation}

The surviving experts keep their original weights, so the expert-branch weight sum decreases to $m_T-m_D$. The residual remains $h$.

Remaining-expert renormalization multiplies the surviving weighted contribution by the gain $A_{\gamma} = 1/(m_T-m_D)$:

\begin{equation}
o_{\gamma} = h + \frac{1}{m_T - m_D} \sum_{i \in T \setminus D} g_i\, y_i .
\end{equation}

Freezing removes the disabled contribution. Renormalization then applies an additional gain to every surviving contribution. The first effect changes the selected content, and the second changes its scale relative to the residual. The weight removed by freezing and the increase in total weight under renormalization are:

\begin{equation}
\resizebox{0.88\columnwidth}{!}{$\begin{split}
\Delta m_{\beta} &= m_D, \\
\Delta m_{\gamma} &= 1 - m_T \;\; \text{(raw)}, \qquad 0 \;\; \text{(renorm)} \qquad \text{from intact}, \\
\Delta m_{\gamma} &= 1 - m_T + m_D \;\; \text{(raw)}, \qquad m_D \;\; \text{(renorm)} \qquad \text{from freeze} .
\end{split}$}
\end{equation}

In a raw-probability model, $\gamma$ increases the retained weight sum from $m_T-m_D$ to one. This introduces additional amplification alongside the loss of disabled contributions. Relative to the intact layer, the weight sum increases by $1-m_T$, and relative to the freezing output it increases by $1-m_T+m_D$. In a normalized-weight model, the sum is one before and after deactivation, and it exceeds the freezing weight sum by $m_D$. The surviving weights nevertheless increase by a factor of $A_{\gamma}=1/(1-m_D)$, changing the individual contributions even though their total weight remains fixed.

Unnormalized re-routing weights the new set $T'$ directly by its raw probabilities. For a surviving member of the original active set in a model trained with normalized weights, the ratio between its new and original weight is:

\begin{equation}
\frac{g_i^{(\delta)}}{g_i^{(\text{train})}} = s_T , \qquad i \in T \setminus D .
\end{equation}

Its weighted contribution is thus reduced to $s_T$ times the trained value. Because $T$ already contains the $k$ highest-probability experts, the raw probability sum $s_{T'}$ of the replacement set cannot exceed $s_T$. In normalized-weight models, $\delta$ removes the gain adjustment used during training and reduces the expert-branch weight sum below one. Raw-probability models continue to use their original weighting rule.

Across the four implementations, $\alpha$ recomputes weights under the training rule and $\beta$ preserves the surviving weights. By contrast, $\gamma$ introduces additional amplification in raw-probability models, and $\delta$ introduces attenuation in normalized-weight models. Both mismatches alter the scale of the expert branch relative to the residual. Prediction 1 covers raw-probability models. On a fixed disabled set, remaining-expert renormalization produces the largest perplexity increase among the four implementations. Prediction 2 covers normalized-weight models. On a fixed disabled set, unnormalized re-routing produces the largest perplexity increase among the four implementations.

The model also covers the number of active experts. For fixed router probabilities, $k_1<k_2$ implies $m_T(k_1)\le m_T(k_2)$ and therefore $A_{\gamma}(k_1)\ge A_{\gamma}(k_2)$, so narrowing the active set monotonically increases the gain introduced by additional renormalization. The width experiment therefore tests a direct monotonic consequence of the weight-sum model.

\section{Methodology}
\label{sec:method}

\subsection{Models}

We evaluate five public instruction-tuned checkpoints from four model families. Table~\ref{tab:models} lists their architectures and training-time weight rules. A base OLMoE checkpoint, with the same architecture and rule as its instruction-tuned counterpart, provides an additional comparison. Qwen3-30B-A3B is evaluated for perplexity at two deactivation proportions; the other five checkpoints undergo both perplexity and task evaluation.

\begin{table*}[t]
\centering
\small
\setlength{\tabcolsep}{6pt}
\begin{tabular}{@{}lrrrl@{}}
\toprule
Model & Experts & $k$ & MoE layers & Training rule \\
\midrule
OLMoE-1B-7B & 64 & 8 & 16 & original probabilities \\
DeepSeek-V2-Lite & 64 & 6 & 26 & original probabilities \\
Granite-3.1-3B & 40 & 8 & 32 & renormalized \\
Granite-3.1-1B & 32 & 8 & 24 & renormalized \\
Qwen3-30B & 128 & 8 & 48 & renormalized \\
\bottomrule
\end{tabular}
\caption{Architectures and training-time weight rules of the five instruction-tuned checkpoints. The additional base OLMoE checkpoint shares its architecture and rule with the instruction-tuned version.}
\label{tab:models}
\end{table*}

\subsection{Disabled Sets}

The calibration corpus comprises the first 2,000 documents from the first C4 shard, packed into 128 sequences of 2,048 tokens \citep{raffel2020t5}. We rank experts by activation and select the lowest-scoring experts separately at each layer. Two additional criteria provide alternative disabled sets: selection frequency, which ranks experts by how often they are chosen, and random sampling with seeds 42, 43, and 44. The deactivation proportions are 12.5\%, 25\%, 37.5\%, and 50\%; Appendix~\ref{app:ppl} lists the evaluated configurations. For a given model, proportion, and selection criterion, the disabled set is identical across all four implementations.

\subsection{Implementation}

All four operations are applied directly to the released checkpoints, with model parameters held fixed. Removal masks disabled experts before the softmax and recomputes the probabilities. Unnormalized re-routing excludes them only during selection, retaining the original softmax denominator. Freezing zeros their contributions after routing, whereas remaining-expert renormalization rescales the surviving weights. The two re-routing implementations fill the vacated slots with replacement experts; the other two retain only the original survivors.

We use Hugging Face Transformers for model loading and the EleutherAI Evaluation Harness for evaluation \citep{wolf2020transformers,gao2024lmeval}. Implementation checks include reproducing the intact model with an empty disabled set and comparing the eager execution path with the fused kernel. Appendix~\ref{app:impl} gives the implementation settings.

\subsection{Evaluation}

We compute WikiText-2 test perplexity using sliding windows of 1,024 tokens and a stride of 512, covering the full test text \citep{merity2017pointer}. Multiple-choice evaluation includes ARC-easy, ARC-challenge, OpenBookQA, HellaSwag, BoolQ, RTE, and WinoGrande \citep{clark2018arc,mihaylov2018obqa,zellers2019hellaswag,clark2019boolq,wang2019glue,sakaguchi2021winogrande}, together with MMLU \citep{hendrycks2021mmlu}. Uniform guessing provides reference accuracies of 25\% for four-option tasks and 50\% for two-option tasks.

Free-form generation is evaluated on GSM8K using strict and flexible matching of the final answer \citep{cobbe2021gsm8k}. We retain item-level likelihoods and generated outputs for paired comparisons and analysis of generation behavior.

\subsection{Statistical Testing}

We use the exact paired McNemar test to compare correctness on identical items, counting changes from correct to incorrect and vice versa between implementations \citep{mcnemar1947}. The correction family contains seven likelihood tasks, namely ARC-easy, ARC-challenge, OpenBookQA, BoolQ, RTE, WinoGrande, and MMLU, crossed with three implementation pairs, namely $\alpha$ against $\beta$ at 25\% deactivation, $\alpha$ against $\beta$ at 50\% deactivation, and $\gamma$ against $\beta$ at 50\% deactivation, for 21 paired tests. We apply the Benjamini-Hochberg correction and report two-sided $q$ values \citep{benjamini1995}.

\section{Results}
\label{sec:results}

\subsection{The Weight-Sum Model Predicts the Highest-Loss Implementation}
\label{sec:ladders}

\begin{figure}[t]
\centering
\includegraphics[width=\columnwidth]{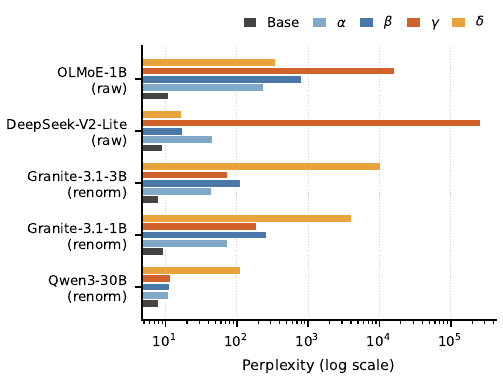}
\caption{Perplexity at 50\% deactivation using activation-based selection, on a log scale. The largest increase occurs under renormalization in raw-probability models and unnormalized re-routing in normalized-weight models.}
\label{fig:ladders}
\end{figure}

Across all five instruction-tuned checkpoints, the weight-sum model identifies the highest-perplexity implementation from the training-time weight rule before any evaluation is run. Raw-probability models peak under $\gamma$, and normalized-weight models peak under $\delta$, as Table~\ref{tab:ppl} and Figure~\ref{fig:ladders} show. Both directional predictions hold at every tested proportion under activation-based selection.

\begin{table*}[t]
\centering
\small
\setlength{\tabcolsep}{6.5pt}
\begin{tabular}{@{}lrrrrr@{}}
\toprule
Model & Base & $\alpha$ & $\beta$ & $\gamma$ & $\delta$ \\
\midrule
OLMoE-1B, raw        & 10.89 & 232.77 & 798.58 & 16,072.70 & 352.42 \\
DeepSeek-V2-Lite, raw & 9.00 & 45.20 & 17.21 & 258,621.64 & 16.64 \\
Granite-3.1-3B, renorm & 7.89 & 44.26 & 112.63 & 73.98 & 10,301.60 \\
Granite-3.1-1B, renorm & 9.22 & 73.28 & 256.08 & 186.57 & 4,095.87 \\
Qwen3-30B, renorm    & 8.01 & 10.90 & 11.23 & 11.54 & 111.84 \\
\bottomrule
\end{tabular}
\caption{Perplexity at 50\% deactivation using activation-based selection. Weight-rule mismatch corresponds to $\gamma$ in raw-probability models and $\delta$ in normalized-weight models. The intact model provides the baseline for each checkpoint.}
\label{tab:ppl}
\end{table*}

On an identical disabled set, $\beta$ and $\gamma$ use the same routing procedure and differ in whether the surviving weights are rescaled. On DeepSeek-V2-Lite, freezing yields a perplexity of 17.21 and renormalization yields 258,621.64, a factor of 15,027 from weight handling alone. This DeepSeek comparison isolates weight handling as a source of model collapse. On OLMoE, the perplexities under $\alpha$ and $\gamma$ differ by a factor of 69.

At 25\% deactivation with activation-based selection, perplexity under the training rule is approximately 8 to 24, whereas mismatched implementations already show substantial increases. Selection also matters within a given implementation. On OLMoE at this proportion, $\gamma$ yields a perplexity of 7,890 with activation-based selection and 7,086 with frequency-based selection; random sets yield 10,203 to 12,873. Tables~\ref{tab:c1}--\ref{tab:c5} in Appendix~\ref{app:ppl} give the full perplexity results for all six checkpoints.

\subsection{Likelihood, Accuracy, and Generation Fail on Distinct Axes}
\label{sec:margin}

Multiple-choice accuracy depends on the relative ranking of candidate answers, whereas perplexity measures sequence likelihood. These metrics capture different aspects of performance after deactivation. Table~\ref{tab:acc} reports OLMoE results at 50\% deactivation, and Tables~\ref{tab:d1}--\ref{tab:d5} in Appendix~\ref{app:tasks} give the complete task results.

Under freezing, OLMoE scores 24.0\% on MMLU, 23.0\% on ARC-challenge, and 19.0\% on OpenBookQA, approaching or falling below the four-option chance level. Its two-option accuracies remain above chance: 58.2\% on BoolQ, 59.6\% on WinoGrande, and 55.2\% on RTE. ARC-easy remains above 25\% when the training rule is retained and reaches 27.7\% under $\gamma$. The same renormalization implementation reduces BoolQ and WinoGrande to 41.7\% and 48.9\%, respectively, while RTE remains at 53.1\%. Its effect therefore varies across tasks.

The paired comparisons also reveal differences between implementations. At 25\% deactivation on MMLU, $\beta$ answers 360 items correctly that $\alpha$ answers incorrectly; the reverse occurs on 215 items, giving an adjusted $q=1.6\times10^{-9}$. At 50\% deactivation, the relevant paired differences on BoolQ, ARC-easy, and WinoGrande remain significant, with $q<10^{-6}$. On MMLU, the two discordant counts for $\gamma$ versus $\beta$ are 313 and 265, with $q=0.067$; both accuracies are near chance.

\begin{table}[t]
\centering
\small
\setlength{\tabcolsep}{4.5pt}
\begin{tabular}{@{}lrrrr@{}}
\toprule
Task & Base & $\alpha$ & $\beta$ & $\gamma$ \\
\midrule
\multicolumn{5}{@{}l}{\textit{Four-option, chance level near 25}} \\
ARC-easy     & 77.3 & 43.3 & 35.8 & 27.7 \\
ARC-challenge & 49.7 & 26.0 & 23.0 & 21.7 \\
OpenBookQA   & 32.8 & 21.2 & 19.0 & 15.2 \\
MMLU         & 54.2 & 24.5 & 24.0 & 26.4 \\
\midrule
\multicolumn{5}{@{}l}{\textit{Two-option, chance level near 50}} \\
BoolQ        & 80.6 & 56.4 & 58.2 & 41.7 \\
RTE          & 67.1 & 53.1 & 55.2 & 53.1 \\
WinoGrande   & 69.0 & 56.7 & 59.6 & 48.9 \\
\midrule
\multicolumn{5}{@{}l}{\textit{Free-form, chance level near 0}} \\
GSM8K        & 36.2 & 0.0 & 0.0 & 0.0 \\
\bottomrule
\end{tabular}
\caption{OLMoE task accuracy at 50\% deactivation, grouped by task type. Four-option and two-option tasks have uniform-guessing references of 25\% and 50\%, respectively. GSM8K uses strict final-answer matching. HellaSwag results are reported in Appendix~\ref{app:tasks} for random configurations and Appendix~\ref{app:width} for width configurations.}
\label{tab:acc}
\end{table}

OLMoE's GSM8K strict-match baseline is 36.2\%. With half of the experts deactivated, strict-match accuracy falls to zero under every implementation evaluated, as Table~\ref{tab:acc} shows. Flexible matching gives 0.8\% for $\alpha$ and $\gamma$, while $\beta$ remains at zero. Generated outputs reveal distinct failure patterns among implementations with the same strict-match accuracy.

DeepSeek-V2-Lite shows a similar separation across evaluation types. Under implementations that retain its training rule, MMLU accuracy ranges from 23.5\% to 25.5\%. Removal gives 56.3\% on BoolQ, 56.3\% on RTE, and 63.2\% on WinoGrande; $\gamma$ reduces BoolQ to 41.7\%. The intact GSM8K strict-match accuracy is 66.2\%, while the highest flexible-match accuracy after deactivation is 1.8\%. Table~\ref{tab:d5} reports the two matching criteria separately.

\subsection{Expert Selection Controls the Magnitude of Degradation}

Within a fixed implementation, the disabled set determines the magnitude of degradation. With half of Granite-3B's experts frozen, activation-based selection yields a perplexity of 112.63, while three random sets yield 11,428 to 93,759, approximately 101 to 832 times the perplexity under activation-based selection. Implementation differences appear in the base checkpoint as well, with $\alpha$ at 26.25 and $\beta$ at 49.66 under 37.5\% random deactivation, as Table~\ref{tab:c1} shows, so the phenomenon does not depend on instruction tuning. The corresponding task results are reported in Appendix~\ref{app:tasks}.

\subsection{The Weight-Sum Model Predicts the Width Trend}
\label{sec:width}

For fixed router probabilities, reducing the number of active experts $k$ lowers the probability sum of the selected set and increases the scaling factor introduced by renormalization. The weight-sum model therefore predicts that additional renormalization in a raw-probability model will become more damaging as $k$ decreases. This extends the prediction from fixed expert deactivation to inference-time width adjustment.

\begin{figure}[t]
\centering
\includegraphics[width=\columnwidth]{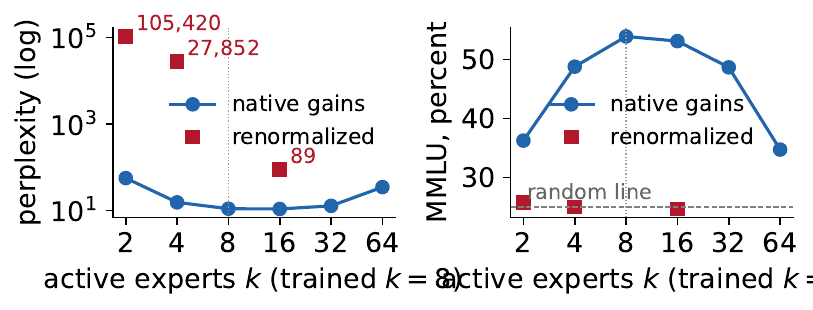}
\caption{OLMoE across active-expert counts. The panels show perplexity on a log scale (left) and MMLU (right); the dotted line marks the training width. Additional renormalization causes greater perplexity degradation at smaller widths.}
\label{fig:width}
\end{figure}

We vary OLMoE's active-expert count from 2 to 64 and compare raw-probability weighting with renormalization at the tested non-training widths (Table~\ref{tab:width} and Figure~\ref{fig:width}). Appendix~\ref{app:width} gives the width configurations for both OLMoE checkpoints.

\begin{table}[t]
\centering
\small
\setlength{\tabcolsep}{4.5pt}
\begin{tabular}{@{}ccccc@{}}
\toprule
$k$ & \multicolumn{2}{c}{training rule} & \multicolumn{2}{c}{renormalization rule} \\
\cmidrule(lr){2-3}\cmidrule(lr){4-5}
 & PPL & MMLU & PPL & MMLU \\
\midrule
2  & 55.66 & 36.3 & 105,420 & 25.7 \\
4  & 15.28 & 48.8 & 27,852 & 25.0 \\
8  & 10.89 & 53.9 & -- & -- \\
16 & 10.75 & 53.1 & 88.76 & 24.7 \\
32 & 12.69 & 48.7 & -- & -- \\
64 & 34.44 & 34.7 & -- & -- \\
\bottomrule
\end{tabular}
\caption{OLMoE perplexity and MMLU at different active-expert counts. The training width is $k=8$. Renormalization is evaluated at the non-training widths shown; its perplexity penalty increases as the active-expert count decreases.}
\label{tab:width}
\end{table}

Under the native rule, widening to $k=16$ leaves perplexity at 10.75 and MMLU at 53.1\%, level with the trained width. At $k=64$, perplexity rises to 34.44 and MMLU falls to 34.7\%. Relative to the native rule at the same $k$, renormalization increases perplexity by factors of approximately 1,894 at $k=2$, 1,823 at $k=4$, and 8.26 at $k=16$. The perplexity penalty increases as the active-expert count decreases, supporting the model's prediction beyond fixed expert deactivation.

\section{Analysis}
\label{sec:analysis}

\subsection{Weight-Rule Mismatch and Contribution Loss}

Freezing changes the layer output by deleting the disabled experts' weighted contributions. A weight-rule mismatch additionally rescales the retained expert branch. In raw-probability models, $\gamma$ increases the weight sum; in normalized-weight models, $\delta$ reduces it. These changes alter the scale of the expert branch relative to the residual and correspond to the high-loss implementations for the two model classes in Table~\ref{tab:ppl}.

On identical disabled sets, the OLMoE perplexities for $\gamma$ and $\alpha$ are approximately 16,072 and 233, a factor of 69 apart. For DeepSeek-V2-Lite, $\gamma$ and $\beta$ give approximately 258,622 and 17, a factor of 15,027 apart. With $\beta$ fixed, the highest Granite-3B random-set perplexity is 832 times the activation-based value. Implementation specifies how routing and weights are handled, while set selection specifies which expert contributions are removed. Both can substantially affect the magnitude of degradation.

\begin{figure}[t]
\centering
\includegraphics[width=0.92\columnwidth]{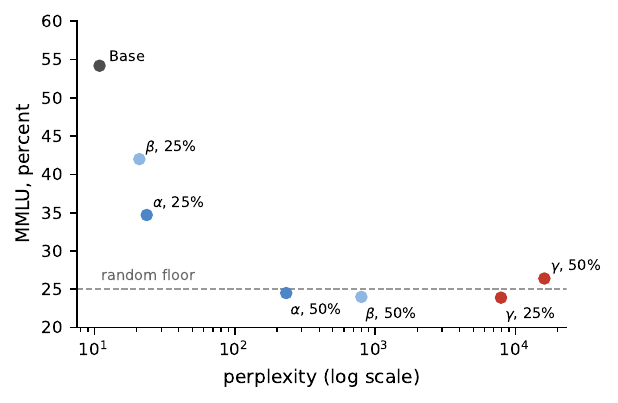}
\caption{Perplexity and MMLU accuracy on OLMoE. Each point is one configuration, with perplexity on a logarithmic horizontal axis. Configurations with a weight-rule mismatch have high perplexity and MMLU accuracy near the chance level.}
\label{fig:dissociation}
\end{figure}

Figure~\ref{fig:dissociation} places OLMoE perplexity and MMLU accuracy on the same plot. At 25\% deactivation, configurations that retain the training rule have relatively low perplexity and high MMLU accuracy. At 50\%, MMLU approaches chance while perplexity still varies considerably. Mismatched configurations occupy the high-perplexity region and also approach chance on MMLU. Configurations with similar answer accuracy can thus retain different sequence likelihoods, making perplexity useful for distinguishing their degree of degradation.

\subsection{Mean Log-Perplexity Increase per Expert Layer}
\label{sec:depth}

Let $L_{\text{MoE}}$ be the number of expert layers and $\pi$ the perplexity ratio between a mismatched configuration and its reference. The width grid uses the width-grid reference, the same checkpoint at the same $k$ under its original rule, and the deactivation grid uses the deactivation-grid reference, the intact baseline of the same checkpoint. We define the mean log increase per expert layer as:

\begin{equation}
\rho = \ln \pi / L_{\text{MoE}} .
\end{equation}

The total log increase can correspondingly be written in terms of the number of expert layers and $\rho$:

\begin{equation}
\ln \pi = L_{\text{MoE}} \, \rho .
\end{equation}

Changing expert weights alters the relative contributions of the expert branch and the residual, and these changes can propagate through subsequent layers. The statistic $\rho$ distributes the final log-perplexity increase over the expert layers, providing a layer-normalized summary of average degradation across configurations.

For the mismatched configurations at 50\% deactivation, DeepSeek-V2-Lite has 26 expert layers and $\rho=0.395$, while OLMoE has 16 layers and $\rho=0.456$. Granite-3B, Granite-1B, and Qwen3 have 32, 24, and 48 expert layers, with $\rho=0.224$, 0.254, and 0.055, respectively. Additional renormalization yields a comparatively large mean increase in both raw-probability models.

In the OLMoE width experiments, $\rho$ is 0.472 at $k=2$ and 0.469 at $k=4$, within 0.016 of the value 0.456 for $\gamma$ at 50\% deactivation. At $k=16$, it decreases to 0.132. The two Granite checkpoints differ by less than 0.03. Narrow-width mismatches and expert deactivation therefore show descriptive agreement across protocols, whereas the increase is smaller after widening, consistent with the direction of the extrapolated prediction.

\subsection{Generation Behavior Across Implementations}
\label{sec:forms}

Even when strict-match accuracy is zero, generated outputs distinguish the failure patterns of different implementations (Appendix~\ref{app:qual}). Outputs under $\alpha$ show prompt echo, those under $\beta$ show token loop, and those under $\gamma$ show early termination. Final-answer accuracy and the form of the generated text therefore provide complementary information about the effects of deactivation.

\section{Related Work}
\label{sec:related}

Expert pruning typically uses importance estimates to select the retained subset and investigates performance recovery after compression. REAP scores experts by router-weighted activation, HOPE propagates dependencies through a sparse graph of expert groups, and Shapley-based pruning estimates each expert's value contribution \citep{lasby2025reap,tseng2026hope,huang2025shapley}. MoNE replaces redundant experts with lighter alternatives, DERN recombines neurons from discarded experts, and EEPA selects subsets by reconstruction error relative to the intact layer output and recomputes weights using the original routing formula \citep{zhang2026mone,zhou2025dern,lu2024eepa}. Alongside subset selection and recovery, the implementation applied to a fixed disabled set directly affects performance.

Studies without recovery training examine fixed-proportion expert removal, subset selection from a few in-domain demonstrations, and dynamic retention of a fraction of the active experts \citep{neogi2025unauthorized,dong2025easyep,chen2026dynamic}. Expert offloading and scheduling further address parameter availability and execution cost \citep{hwang2024pregated,kamahori2025fiddler}. SEUF concentrates unlearning on experts that carry targeted knowledge and stabilizes their routing during training \citep{zhuang2025seuf}. These interventions affect expert availability, selection, or training; the weighting computation determines how the selected expert outputs combine in each layer.

\citet{chen2026halving} reduce the number of experts computed for each token while retaining a normalization reference based on the training-time active set. Their expert bank remains intact, and the router selects experts for each token. Our analysis starts from a fixed disabled set, compares four implementations under two training-time weight rules, and extends the resulting prediction to changes in the active-expert count.

\citet{specialization2024} measure specialization through an expert's influence on downstream routing distributions, compare removal with freezing, and use this comparison to choose a pruning implementation. We incorporate the training-time weight rule into this comparison and cover all four implementations. Knowledge localization, multilingual routing, and post-training steering examine the knowledge carried by experts, language-dependent expert selection, and controlled activation, respectively \citep{bandarkar2026localization,bandarkar2026multilingual,fayyaz2026steermoe}. Elastic MoE and Matryoshka models further investigate variable active-expert counts \citep{gu2025emoe,wang2025matryoshka}.

Sparse gating established the basic formulation of token-dependent expert selection and weighting \citep{shazeer2017outrageously,fedus2022switch}, and subsequent work has extended expert architectures and routing designs \citep{jiang2024mixtral,dai2024deepseekmoe}. \citet{honda2024aggregate} address shortcut shifts through expert aggregation and post-hoc control of routing. Using the training-time weighting computation as a reference, we explain performance differences among implementations on fixed disabled sets and test the resulting prediction under changes in inference-time width.

These studies establish the importance of expert selection and routing. The present work adds the implementation axis. On identical disabled sets, changing the weight computation alone changes the outcome by orders of magnitude, and the weight-sum model predicts which change is most damaging.

\section{Conclusion}
\label{sec:conclusion}

Expert deactivation is governed by two choices, namely which experts are disabled and how the surviving experts are weighted. The weight-sum model separates the two and turns the training-time weight rule into a predictor of the highest-loss implementation. Five instruction-tuned checkpoints from four families confirm the prediction, and the width results confirm its monotonic extension to active-expert adjustment.

The OLMoE width experiments validate a further prediction derived from fixed expert deactivation: additional renormalization causes more severe degradation when fewer experts are active. The mean log-perplexity increase per expert layer summarizes the magnitude of degradation, while multiple-choice and free-form evaluations reveal different task outcomes. Deployment follows three steps.

\begin{enumerate}[leftmargin=1.4em,topsep=2pt,itemsep=1pt]

\item Read the training-time weight convention first, and determine whether the checkpoint weights selected experts by raw router probabilities or by renormalized selected-set weights.

\item Match the implementation to that convention, re-routing with the native weight rule or preserving surviving weights when slots stay unfilled, and apply additional normalization only where it matches the training computation.

\item Control selection and implementation separately, measure perplexity and task accuracy together, and inspect generated outputs when final-answer scores saturate.

\end{enumerate}

Reproducibility. The study uses public checkpoints, public datasets, documented deactivation rules, and public evaluation frameworks. Section~\ref{sec:background} specifies the routing and weighting operations, and the appendices provide the calibration protocol, evaluation settings, and complete result grids for reproducing the reported measurements.

\section{Limitations}

The evidence covers six checkpoints from four families spanning 1B to 30B parameters, with expert rankings calibrated on a fixed C4 corpus and free-form generation evaluated on GSM8K. Extending model scale, calibration distributions, and generation tasks is the direct next test of the correspondence between training-time weight rules and high-loss implementations. Recovery training after a rule mismatch is unexplored.

\bibliography{references_v0.2.2}
\bibliographystyle{acl_natbib}

\appendix

\clearpage
\section{Implementation Details}
\label{app:impl}

We implement all four implementations as routing operations on the released checkpoints. We use HuggingFace Transformers \citep{wolf2020transformers} to load models and datasets, and the EleutherAI Evaluation Harness \citep{gao2024lmeval} for benchmark evaluation. The protocol covers the nine items below.

\begin{enumerate}[leftmargin=1.4em,topsep=2pt,itemsep=1pt]

\item Checkpoints. Every run pins the released checkpoints named in Table~\ref{tab:models}, and every measured path checks equivalence with the official forward pass.

\item Precision. Every run records its numerical precision, fp16 on the local GPU, bf16 on the rented GPUs, and fp32 under the recompute rule in item 9.

\item Expert scores. Activation and frequency scores are computed on the calibration corpus layer by layer, and experts are ranked within each layer, with the lowest-scoring experts forming the disabled set of that layer.

\item Rounding. The per-layer disabled count is $N$ times the deactivation proportion rounded to the nearest integer, and every tested proportion divides evenly at every layer width.

\item Random sets. The random sets draw the per-layer disabled count with an independent per-layer sampling stream derived from the seeds 42, 43, and 44.

\item Perplexity. WikiText-2 test perplexity uses sliding windows of 1,024 tokens with a stride of 512 over the full test text.

\item Generation. GSM8K generation uses 1,000 test problems for Granite-3B and 500 for OLMoE-Instruct and DeepSeek-V2-Lite. Strict and flexible matchers score the final answer, and item-level likelihoods and generated outputs are retained.

\item Equivalence. The eager execution path is compared with the fused kernel on every measured path.

\item fp32 recompute. A configuration whose fp16 perplexity computation overflows is recomputed in fp32, and its task accuracies remain in fp16.

\end{enumerate}

\renewcommand{\arraystretch}{0.92}

\section{Full Perplexity Grids}
\label{app:ppl}

\begin{table*}[t]
\centering
\scriptsize
\setlength{\tabcolsep}{3pt}
\begin{tabular*}{0.48\textwidth}[t]{@{\extracolsep{\fill}}lrr@{}}
\toprule
\multicolumn{3}{@{}l}{\normalsize OLMoE-1B-7B-0924-Instruct} \\
\midrule
Configuration & PPL & ratio \\
\midrule
base & 10.89 & 1.00 \\
$\alpha$-ean25 & 23.60 & 2.17 \\
$\alpha$-ean50 & 232.77 & 21.4 \\
$\alpha$-rand25-s42 & 18.56 & 1.70 \\
$\alpha$-rand25-s43 & 18.60 & 1.71 \\
$\alpha$-rand25-s44 & 18.28 & 1.68 \\
$\alpha$-rand50-s42 & 106.27 & 9.76 \\
$\alpha$-rand50-s43 & 85.59 & 7.86 \\
$\alpha$-rand50-s44 & 86.00 & 7.90 \\
$\beta$-ean25 & 20.92 & 1.92 \\
$\beta$-ean50 & 798.58 & 73.3 \\
$\beta$-rand25-s42 & 18.06 & 1.66 \\
$\beta$-rand25-s43 & 18.99 & 1.74 \\
$\beta$-rand25-s44 & 18.03 & 1.66 \\
$\beta$-rand50-s42 & 1{,}551.44 & 142 \\
$\beta$-rand50-s43 & 102.50 & 9.41 \\
$\beta$-rand50-s44 & 786.75 & 72.2 \\
$\gamma$-ean25 & 7{,}889.82 & 725 \\
$\gamma$-ean50 & 16{,}072.70 & 1,476 \\
$\gamma$-ean12.5 & 3{,}708.37 & 341 \\
$\gamma$-ean37.5 & 10{,}953.59 & 1,006 \\
$\delta$-ean25 & 19.61 & 1.80 \\
$\delta$-ean50 & 352.42 & 32.4 \\
ke (dense) & 34.44 & 3.16 \\
\midrule
$\alpha$-ean50-ppl & 232.99 & 21.4 \\
$\beta$-ean50-ppl & 797.92 & 73.3 \\
base-ppl & 10.89 & 1.00 \\
$\gamma$-ean50-ppl & 16{,}063.78 & 1,475 \\
$\gamma$-freq25 & 7{,}085.67 & 651 \\
$\gamma$-freq50 & 16{,}084.72 & 1,477 \\
$\gamma$-rand25-s42 & 10{,}203.82 & 937 \\
$\gamma$-rand25-s43 & 10{,}232.50 & 940 \\
$\gamma$-rand25-s44 & 12{,}872.87 & 1,182 \\
$\gamma$-rand50-s42 & 37{,}592.51 & 3,452 \\
$\gamma$-rand50-s43 & 53{,}324.82 & 4,897 \\
$\gamma$-rand50-s44 & 38{,}632.81 & 3,548 \\
\bottomrule
\end{tabular*}\hfill
\begin{tabular*}{0.48\textwidth}[t]{@{\extracolsep{\fill}}lrr@{}}
\toprule
\multicolumn{3}{@{}l}{\normalsize OLMoE-1B-7B-0924-Base} \\
\midrule
Configuration & PPL & ratio \\
\midrule
base & 8.38 & 1.00 \\
$\alpha$-rand25-s42 & 14.99 & 1.79 \\
$\alpha$-rand25-s43 & 14.52 & 1.73 \\
$\alpha$-rand25-s44 & 14.43 & 1.72 \\
$\alpha$-random37.5-s42 & 26.25 & 3.13 \\
$\beta$-rand25-s42 & 14.56 & 1.74 \\
$\beta$-rand25-s43 & 14.21 & 1.70 \\
$\beta$-rand25-s44 & 13.85 & 1.65 \\
$\beta$-random37.5-s42 & 49.66 & 5.93 \\
ke (dense) & 46.14 & 5.51 \\
\bottomrule
\end{tabular*}
\caption{Full perplexity grids for the two OLMoE checkpoints. The Instruct grid lists both disabled shares, the three criteria with seeds, the two intermediate $\gamma$ shares, the independent replicates, and the dense endpoint. The prefixes rand and random denote the same random-sampling criterion. Each grid uses its own paired reference run for within-grid ratios, and the perplexity ratio is the value over the base row of the same table.}
\label{tab:c1}
\end{table*}

\begin{table*}[t]
\centering
\scriptsize
\setlength{\tabcolsep}{3pt}
\begin{tabular*}{0.48\textwidth}[t]{@{\extracolsep{\fill}}lrr@{}}
\toprule
\multicolumn{3}{@{}l}{\normalsize Granite-3.1-3B-A800M} \\
\midrule
Configuration & PPL & ratio \\
\midrule
base & 7.89 & 1.00 \\
ke (dense) & 39.05 & 4.95 \\
$\alpha$-ean25 & 12.85 & 1.63 \\
$\alpha$-ean50 & 44.26 & 5.61 \\
$\alpha$-rand25-s42 & 745.75 & 94.5 \\
$\alpha$-rand25-s43 & 24.82 & 3.15 \\
$\alpha$-rand25-s44 & 20.76 & 2.63 \\
$\alpha$-rand50-s42 & 283.79 & 36.0 \\
$\alpha$-rand50-s43 & 285.94 & 36.2 \\
$\alpha$-rand50-s44 & 1{,}311.94 & 166 \\
$\beta$-ean25 & 16.23 & 2.06 \\
$\beta$-ean50 & 112.63 & 14.3 \\
$\beta$-rand25-s42 & 783.60 & 99.3 \\
$\beta$-rand25-s43 & 69.92 & 8.86 \\
$\beta$-rand25-s44 & 85.91 & 10.9 \\
$\beta$-rand50-s42 & 11{,}428.03 & 1,448 \\
$\beta$-rand50-s43 & 11{,}592.18 & 1,469 \\
$\beta$-rand50-s44 & 93{,}759.23 & 11,883 \\
$\gamma$-ean25 & 13.37 & 1.69 \\
$\gamma$-ean50 & 73.98 & 9.38 \\
$\gamma$-rand50-s42 & 1{,}054.09 & 134 \\
$\gamma$-rand50-s43 & 2{,}213.97 & 281 \\
$\gamma$-rand50-s44 & 3{,}071.37 & 389 \\
$\delta$-ean25 & 1{,}050.56 & 133 \\
$\delta$-ean50 & 10{,}301.60 & 1,306 \\
\bottomrule
\end{tabular*}\hfill
\begin{tabular*}{0.48\textwidth}[t]{@{\extracolsep{\fill}}lrr@{}}
\toprule
\multicolumn{3}{@{}l}{\normalsize Granite-3.1-1B-A400M} \\
\midrule
Configuration & PPL & ratio \\
\midrule
base & 9.22 & 1.00 \\
ke (dense) & 23.43 & 2.54 \\
$\alpha$-ean25 & 14.78 & 1.60 \\
$\alpha$-ean50 & 73.28 & 7.95 \\
$\alpha$-rand25-s42 & 27.21 & 2.95 \\
$\alpha$-rand25-s43 & 21.84 & 2.37 \\
$\alpha$-rand25-s44 & 29.31 & 3.18 \\
$\alpha$-rand50-s42 & 5{,}210.22 & 565 \\
$\alpha$-rand50-s43 & 320.15 & 34.7 \\
$\alpha$-rand50-s44 & 1{,}289.51 & 140 \\
$\beta$-ean25 & 17.73 & 1.92 \\
$\beta$-ean50 & 256.08 & 27.8 \\
$\beta$-rand25-s42 & 54.45 & 5.91 \\
$\beta$-rand25-s43$^\dag$ & 29.12$^\dag$ & 3.16 \\
$\beta$-rand25-s44 & 101.37 & 11.0 \\
$\beta$-rand50-s42 & 10{,}548.31 & 1,144 \\
$\beta$-rand50-s43$^\dag$ & 3{,}532.66$^\dag$ & 383 \\
$\beta$-rand50-s44 & 3{,}190.57 & 346 \\
$\gamma$-ean25 & 15.96 & 1.73 \\
$\gamma$-ean50 & 186.57 & 20.2 \\
$\gamma$-rand50-s42 & 10{,}823.05 & 1,174 \\
$\gamma$-rand50-s43 & 1{,}919.55 & 208 \\
$\gamma$-rand50-s44 & 3{,}104.71 & 337 \\
$\delta$-ean25 & 294.81 & 32.0 \\
$\delta$-ean50 & 4{,}095.87 & 444 \\
\bottomrule
\end{tabular*}
\caption{Full perplexity grids for the two Granite checkpoints. The prefixes rand and random denote the same random-sampling criterion. Each grid uses its own paired reference run for within-grid ratios, and the perplexity ratio is the value over the base row of the same table. $^\dag$ marks the configurations recomputed in fp32 under the trigger rule in Appendix~\ref{app:impl}, with task accuracies of those runs in fp16 throughout.}
\label{tab:c3}
\end{table*}

\begin{table*}[t]
\centering
\scriptsize
\setlength{\tabcolsep}{3pt}
\begin{tabular*}{0.48\textwidth}[t]{@{\extracolsep{\fill}}lrr@{}}
\toprule
\multicolumn{3}{@{}l}{\normalsize DeepSeek-V2-Lite (non-renormalizing)} \\
\midrule
Configuration & PPL & ratio \\
\midrule
base & 9.00 & 1.00 \\
$\alpha$-ean25 & 11.57 & 1.29 \\
$\alpha$-ean50 & 45.20 & 5.02 \\
$\beta$-ean25 & 10.45 & 1.16 \\
$\beta$-ean50 & 17.21 & 1.91 \\
$\gamma$-ean25 & 5{,}310.41 & 590 \\
$\gamma$-ean50 & 258{,}621.64 & 28{,}739 \\
$\gamma$-ean12.5 & 354.36 & 39 \\
$\gamma$-ean37.5 & 53{,}154.73 & 5{,}907 \\
$\delta$-ean25 & 10.34 & 1.15 \\
$\delta$-ean50 & 16.64 & 1.85 \\
\bottomrule
\end{tabular*}\hfill
\begin{tabular*}{0.48\textwidth}[t]{@{\extracolsep{\fill}}lrr@{}}
\toprule
\multicolumn{3}{@{}l}{\normalsize Qwen3-30B-A3B-2507 (renormalizing)} \\
\midrule
Configuration & PPL & ratio \\
\midrule
base & 8.01 & 1.00 \\
$\alpha$-ean25 & 8.26 & 1.03 \\
$\alpha$-ean50 & 10.90 & 1.36 \\
$\beta$-ean25 & 8.24 & 1.03 \\
$\beta$-ean50 & 11.23 & 1.40 \\
$\gamma$-ean25 & 8.31 & 1.04 \\
$\gamma$-ean50 & 11.54 & 1.44 \\
$\delta$-ean25 & 28.38 & 3.54 \\
$\delta$-ean50 & 111.84 & 13.96 \\
\bottomrule
\end{tabular*}
\caption{Full perplexity grids for the two cross-architecture replicates. The DeepSeek grid covers two disabled shares and the two intermediate $\gamma$ shares, and the Qwen3 grid covers two disabled shares. The perplexity ratio is the value over the base row of the same table, and each grid uses its own paired reference run for within-grid ratios.}
\label{tab:c5}
\end{table*}

\section{Full Task Grids}
\label{app:tasks}

\begin{table*}[t]
\centering\scriptsize\setlength{\tabcolsep}{3pt}
\caption{Task accuracies for OLMoE-1B-7B-0924-Instruct on the calibrated, frequency, and random sets, the task side of the perplexity grid in Table~\ref{tab:c1}.  A blank marks a task not run for that configuration.}
\label{tab:d1}
\begin{tabular}{@{}lrrrrrrrr@{}}
\toprule
Configuration & ARC-e & ARC-c & BoolQ & OBQA & RTE & Wino & MMLU & HSwg \\
\midrule
base & 77.3 & 49.7 & 80.6 & 32.8 & 67.1 & 69.0 & 54.2 &  \\
$\alpha$-ean25 & 67.7 & 41.9 & 75.5 & 30.4 & 68.6 & 67.3 & 34.7 &  \\
$\alpha$-ean50 & 43.3 & 26.0 & 56.4 & 21.2 & 53.1 & 56.7 & 24.5 &  \\
$\alpha$-rand25-s42 & 68.4 & 41.8 & 73.8 & 28.4 & 68.6 & 61.6 & 39.9 & 50.5 \\
$\alpha$-rand25-s43 & 71.5 & 40.6 & 70.2 & 29.8 & 58.5 & 63.1 & 43.6 & 50.9 \\
$\alpha$-rand25-s44 & 70.3 & 40.9 & 72.9 & 31.6 & 69.7 & 63.2 & 40.9 & 51.1 \\
$\alpha$-rand50-s42 & 34.8 & 21.2 & 57.3 & 17.2 & 54.5 & 51.5 & 27.4 &  \\
$\beta$-ean25 & 68.5 & 42.9 & 78.2 & 31.0 & 72.2 & 68.2 & 42.0 &  \\
$\beta$-ean50 & 35.8 & 23.0 & 58.2 & 19.0 & 55.2 & 59.6 & 24.0 &  \\
$\beta$-rand25-s42 & 70.0 & 43.3 & 75.9 & 27.0 & 70.0 & 62.9 & 42.6 & 51.8 \\
$\beta$-rand25-s43 & 73.3 & 43.0 & 72.0 & 31.2 & 59.6 & 62.7 & 40.9 & 51.9 \\
$\beta$-rand25-s44 & 72.1 & 43.9 & 76.2 & 30.4 & 67.9 & 64.7 & 42.3 & 51.9 \\
$\beta$-rand50-s42 & 30.9 & 21.1 & 63.3 & 12.8 & 51.6 & 52.1 & 28.4 &  \\
$\gamma$-ean25 & 27.1 & 20.1 & 42.8 & 14.8 & 54.2 & 48.8 & 23.9 &  \\
$\gamma$-ean50 & 27.7 & 21.7 & 41.7 & 15.2 & 53.1 & 48.9 & 26.4 &  \\
ke (dense) & 70.4 & 38.7 & 69.3 & 29.0 & 53.1 & 61.5 & 34.9 &  \\
\midrule
$\gamma$-freq25 & 26.7 & 18.7 & 44.0 & 15.4 & 46.2 & 48.1 & 23.2 & 26.2 \\
$\gamma$-freq50 & 26.0 & 20.0 & 40.3 & 14.0 & 53.8 & 48.8 & 25.1 & 25.7 \\
$\gamma$-rand25-s42 & 28.0 & 21.4 & 46.8 & 15.2 & 48.7 & 49.0 & 24.9 & 26.1 \\
$\gamma$-rand25-s43 & 28.3 & 20.6 & 43.7 & 14.6 & 45.5 & 50.8 & 25.0 & 25.7 \\
$\gamma$-rand25-s44 & 27.4 & 21.8 & 46.5 & 12.6 & 54.2 & 49.4 & 24.6 & 25.8 \\
$\gamma$-rand50-s42 & 25.1 & 20.6 & 45.7 & 14.4 & 49.1 & 49.0 & 26.7 & 26.2 \\
$\gamma$-rand50-s43 & 26.5 & 23.8 & 44.3 & 16.8 & 54.5 & 49.9 & 25.3 & 25.7 \\
$\gamma$-rand50-s44 & 26.8 & 20.9 & 43.8 & 15.0 & 47.7 & 49.9 & 24.8 & 25.7 \\
\bottomrule
\end{tabular}
\end{table*}

\begin{table*}[t]
\centering\scriptsize\setlength{\tabcolsep}{3pt}
\caption{Task accuracies for Granite-3.1-3B-A800M, the task side of the perplexity grid in Table~\ref{tab:c3}. The GSM8K column is strict and flexible accuracy in percent on 1{,}000 problems.}
\label{tab:d3}
\begin{tabular}{@{}lrrrrrrrrr@{}}
\toprule
Configuration & ARC-e & ARC-c & BoolQ & OBQA & RTE & Wino & MMLU & HSwg & GSM8K \\
\midrule
base & 74.6 & 43.9 & 79.2 & 31.8 & 73.6 & 67.7 & 52.4 &  & 55.3, 56.8 \\
ke (dense) & 50.0 & 25.0 & 65.9 & 15.6 & 61.0 & 53.0 & 29.9 &  &  \\
$\alpha$-ean25 & 63.6 & 36.3 & 77.3 & 24.0 & 70.0 & 63.5 & 42.2 &  &  \\
$\alpha$-ean50 & 44.5 & 25.3 & 67.7 & 19.0 & 63.9 & 57.6 & 30.4 &  & 0.0, 0.6 \\
$\alpha$-rand25-s42 & 26.3 & 21.1 & 60.8 & 13.0 & 52.0 & 49.2 & 26.9 &  &  \\
$\alpha$-rand25-s43 & 59.5 & 32.4 & 69.1 & 23.4 & 63.9 & 58.5 & 31.9 &  &  \\
$\alpha$-rand25-s44 & 54.3 & 28.6 & 66.5 & 17.8 & 50.2 & 56.7 & 30.2 &  &  \\
$\alpha$-rand50-s42 & 30.8 & 21.2 & 60.3 & 11.2 & 54.9 & 49.7 & 24.0 &  &  \\
$\alpha$-rand50-s43 & 30.8 & 20.7 & 48.9 & 14.0 & 50.5 & 50.8 & 24.4 &  &  \\
$\alpha$-rand50-s44 & 26.6 & 20.0 & 54.6 & 13.8 & 53.1 & 48.1 & 26.2 &  &  \\
$\beta$-ean25 & 60.4 & 34.1 & 74.8 & 25.0 & 74.4 & 62.3 & 41.4 &  &  \\
$\beta$-ean50 & 38.6 & 24.0 & 50.1 & 16.0 & 61.7 & 52.8 & 25.6 &  & 0.0, 0.0 \\
$\beta$-rand25-s42 & 25.9 & 21.4 & 61.4 & 16.0 & 53.4 & 47.6 & 23.5 &  &  \\
$\beta$-rand25-s43 & 50.5 & 25.5 & 64.2 & 19.2 & 58.1 & 53.4 & 23.8 &  &  \\
$\beta$-rand25-s44 & 39.9 & 24.3 & 50.9 & 16.0 & 48.4 & 53.2 & 24.8 &  &  \\
$\beta$-rand50-s42 & 27.9 & 21.6 & 58.4 & 14.2 & 49.5 & 49.3 & 24.1 &  &  \\
$\beta$-rand50-s43 & 27.3 & 21.3 & 53.9 & 15.6 & 53.8 & 49.2 & 24.3 &  &  \\
$\beta$-rand50-s44 & 26.0 & 21.9 & 39.2 & 13.8 & 49.5 & 50.4 & 24.2 &  &  \\
$\gamma$-ean25 & 61.3 & 35.1 & 75.0 & 24.8 & 67.1 & 62.8 & 38.4 &  &  \\
$\gamma$-ean50 & 39.6 & 25.8 & 61.6 & 18.4 & 61.0 & 54.6 & 27.3 &  & 0.0, 0.7 \\
$\delta$-ean50 & 24.7 & 22.3 & 54.7 & 14.4 & 53.1 & 48.9 & 23.8 &  & 0.0, 0.0 \\
\bottomrule
\end{tabular}
\end{table*}

\begin{table*}[t]
\centering\scriptsize\setlength{\tabcolsep}{3pt}
\caption{Task accuracies for Granite-3.1-1B-A400M, the task side of the perplexity grid in Table~\ref{tab:c3}.}
\label{tab:d4}
\begin{tabular}{@{}lrrrrrrrr@{}}
\toprule
Configuration & ARC-e & ARC-c & BoolQ & OBQA & RTE & Wino & MMLU & HSwg \\
\midrule
base & 69.5 & 33.4 & 69.3 & 27.6 & 65.0 & 60.6 & 31.4 &  \\
ke (dense) & 55.2 & 24.3 & 64.2 & 16.0 & 52.7 & 56.7 & 28.1 &  \\
$\alpha$-ean25 & 57.2 & 28.9 & 61.0 & 22.4 & 59.2 & 60.9 & 27.4 &  \\
$\alpha$-ean50 & 36.2 & 22.7 & 56.2 & 17.2 & 53.1 & 53.3 & 24.1 &  \\
$\alpha$-rand25-s42 & 45.1 & 24.0 & 62.8 & 15.8 & 53.8 & 54.1 & 26.6 &  \\
$\alpha$-rand25-s43 & 55.0 & 25.9 & 57.8 & 19.0 & 54.2 & 54.3 & 25.5 &  \\
$\alpha$-rand25-s44 & 50.5 & 24.8 & 61.9 & 15.8 & 52.3 & 52.6 & 25.6 &  \\
$\alpha$-rand50-s42 & 26.1 & 20.7 & 41.6 & 13.8 & 45.8 & 50.4 & 23.9 &  \\
$\alpha$-rand50-s43 & 28.5 & 18.5 & 50.4 & 15.0 & 53.1 & 48.7 & 24.3 &  \\
$\alpha$-rand50-s44 & 26.6 & 20.6 & 41.3 & 12.8 & 49.8 & 49.2 & 24.0 &  \\
$\beta$-ean25 & 56.1 & 27.8 & 61.3 & 20.2 & 57.8 & 59.8 & 25.2 &  \\
$\beta$-ean50 & 35.1 & 21.2 & 49.7 & 15.8 & 52.0 & 51.8 & 23.4 &  \\
$\beta$-rand25-s42 & 40.4 & 21.0 & 57.1 & 14.6 & 54.9 & 53.4 & 24.5 &  \\
$\beta$-rand25-s43$^\dag$ & 53.6 & 25.2 & 54.1 & 17.8 & 47.3 & 53.4 & 25.3 &  \\
$\beta$-rand25-s44 & 41.0 & 21.2 & 55.4 & 14.6 & 51.6 & 50.2 & 26.7 &  \\
$\beta$-rand50-s42 & 26.2 & 21.4 & 38.1 & 13.8 & 45.8 & 50.4 & 24.4 &  \\
$\beta$-rand50-s43$^\dag$ & 25.6 & 22.3 & 52.9 & 16.6 & 51.6 & 48.0 & 24.1 &  \\
$\beta$-rand50-s44 & 27.3 & 21.1 & 38.5 & 13.2 & 48.0 & 49.7 & 24.7 &  \\
$\gamma$-ean25 & 54.7 & 28.3 & 56.4 & 22.4 & 61.0 & 60.8 & 27.9 &  \\
$\gamma$-ean50 & 32.7 & 23.2 & 49.1 & 17.0 & 55.6 & 50.2 & 24.1 &  \\
$\delta$-ean50 & 27.4 & 21.8 & 41.3 & 12.8 & 52.7 & 51.5 & 23.3 &  \\
\bottomrule
\end{tabular}
\end{table*}

\begin{table*}[t]
\centering
\scriptsize
\setlength{\tabcolsep}{3pt}
\begin{tabular}{@{}lrrrrrrrrr@{}}
\toprule
\multicolumn{10}{@{}l}{\normalsize OLMoE-1B-7B-0924-Base} \\
\midrule
Configuration & ARC-e & ARC-c & BoolQ & OBQA & RTE & Wino & MMLU & HSwg & GSM8K \\
\midrule
base & 78.4 & 46.8 & 74.6 & 32.6 & 54.2 & 69.4 & 51.8 & 57.9 &  \\
$\alpha$-rand25-s42 & 67.8 & 36.7 & 67.7 & 27.4 & 49.1 & 64.4 & 34.2 & 48.2 &  \\
$\alpha$-rand25-s43 & 72.4 & 40.0 & 68.5 & 26.0 & 47.3 & 63.4 & 39.9 & 49.1 &  \\
$\alpha$-rand25-s44 & 71.6 & 39.1 & 69.3 & 29.0 & 50.5 & 62.7 & 37.5 & 48.8 &  \\
$\alpha$-random37.5-s42 & 49.7 & 24.2 & 57.1 & 20.2 & 50.2 & 57.9 & 24.6 & 40.2 &  \\
$\beta$-rand25-s42 & 68.4 & 38.0 & 69.5 & 27.4 & 49.1 & 62.0 & 37.7 & 49.5 &  \\
$\beta$-rand25-s43 & 73.6 & 40.7 & 69.0 & 25.6 & 44.8 & 64.2 & 37.9 & 49.8 &  \\
$\beta$-rand25-s44 & 71.6 & 39.1 & 71.5 & 27.6 & 55.6 & 63.3 & 38.0 & 49.9 &  \\
$\beta$-random37.5-s42 & 39.0 & 20.3 & 63.0 & 16.4 & 49.5 & 53.6 & 24.5 & 35.5 &  \\
ke (dense) & 66.1 & 29.6 & 68.0 & 25.4 & 53.1 & 61.3 & 27.1 & 35.8 &  \\
\midrule
\multicolumn{10}{@{}l}{\normalsize DeepSeek-V2-Lite (non-renormalizing)} \\
\midrule
base & 80.3 & 52.6 & 83.1 & 35.2 & 72.9 & 71.7 & 58.5 &  & 66.2, 67.2 \\
$\alpha$-ean50 & 52.7 & 30.7 & 56.3 & 26.2 & 56.3 & 63.2 & 23.5 &  & 0.0, 1.4 \\
$\beta$-ean50 & 64.4 & 35.3 & 78.7 & 26.4 & 63.9 & 69.1 & 25.0 &  & 0.4, 1.0 \\
$\gamma$-ean50 & 26.3 & 20.6 & 41.7 & 17.0 & 49.1 & 51.4 & 25.2 &  & 0.0, 0.6 \\
$\delta$-ean50 & 67.1 & 38.0 & 79.1 & 26.8 & 67.9 & 72.2 & 25.5 &  & 0.8, 1.8 \\
\bottomrule
\end{tabular}
\caption{Task accuracies for OLMoE-1B-7B-0924-Base, which completes the OLMoE family grid of Table~\ref{tab:d1}, and for DeepSeek-V2-Lite (non-renormalizing), the task side of the cross-architecture pair in Table~\ref{tab:c5}.  The GSM8K column reports strict and flexible accuracy in percent on 500 problems for DeepSeek-V2-Lite.}
\label{tab:d5}
\end{table*}

\section{Generation Behavior across Deactivation Implementations}
\label{app:qual}

With half of its experts deactivated, OLMoE exhibits prompt echo, token loops, and early termination in the outputs below.  We quote the start of each output on the first item of the GSM8K test set, which asks the following question.

\begin{quote}
\small Janet's ducks lay 16 eggs per day.  She eats three for breakfast every morning and bakes muffins for her friends every day with four.  She sells the remainder at the farmers' market daily for \$2 per fresh duck egg.  How much in dollars does she make every day at the farmers' market.
\end{quote}

\paragraph{Removal.}  The output shows prompt echo and repeats the following.
\begin{quote}
\small\ttfamily The duck lays 16 eggs a week. The duck is a fresh egg. The duck is a fresh egg is a fresh egg is a fresh is a fresh is a fresh is a fresh is a fresh is a \$ of the fresh, and so on.
\end{quote}
\begin{quote}
\small\ttfamily He runs a week. He runs a week. He runs a week. He runs a week. He runs a week. He runs a week., and so on.
\end{quote}

\paragraph{Freezing.}  The output shows token loop.
\begin{quote}
\small\ttfamily that in- in- that, that, that, that, that, that, that,, and so on.
\end{quote}
\begin{quote}
\small\ttfamily that will, in the question, that will, that that, that that, that that, that that, that that,, and so on.
\end{quote}

\paragraph{Renormalization.}  The output shows early termination.
\begin{quote}
\small\ttfamily Australia
\end{quote}
\begin{quote}
\small\ttfamily for the 5- the leader.
\end{quote}

\section{Width Axis Detail}
\label{app:width}

\begin{table*}[t]
\centering
\scriptsize
\setlength{\tabcolsep}{3pt}
\begin{tabular}{@{}lrrrrrrrrr@{}}
\toprule
\multicolumn{10}{@{}l}{\normalsize OLMoE-1B-7B-0924-Instruct} \\
\midrule
Configuration & PPL & ARC-e & ARC-c & BoolQ & OBQA & RTE & Wino & MMLU & HSwg \\
\midrule
base & 10.89 & 77.4 & 49.2 & 80.6 & 33.0 & 67.5 & 68.7 & 53.9 &  \\
wide-k2 & 55.66 & 60.6 & 31.7 & 65.6 & 19.4 & 53.4 & 57.5 & 36.3 & 42.2 \\
wide-k2-rn & 105{,}420 & 23.2 & 20.6 & 44.0 & 14.8 & 49.1 & 49.2 & 25.7 & 25.5 \\
wide-k4 & 15.28 & 69.8 & 44.3 & 75.7 & 33.6 & 62.1 & 62.6 & 48.8 & 57.2 \\
wide-k4-rn & 27{,}852 & 24.5 & 20.3 & 44.6 & 14.8 & 53.1 & 52.6 & 25.0 & 25.6 \\
wide-k16 & 10.75 & 78.1 & 49.2 & 78.1 & 33.6 & 67.9 & 69.8 & 53.1 &  \\
wide-k16-rn & 88.76 & 54.2 & 25.9 & 49.4 & 21.8 & 54.5 & 50.6 & 24.7 &  \\
wide-k32 & 12.69 & 76.5 & 48.2 & 78.2 & 34.6 & 69.7 & 69.0 & 48.7 &  \\
ke (dense) & 34.44 & 70.4 & 38.7 & 69.2 & 29.0 & 53.1 & 61.3 & 34.7 &  \\
\midrule
\multicolumn{10}{@{}l}{\normalsize OLMoE-1B-7B-0924-Base} \\
\midrule
wide-k16 & 8.92 & 78.0 & 44.8 & 74.1 & 31.6 & 61.7 & 69.6 & 51.8 & 55.8 \\
wide-k16-rn & 70.81 & 58.2 & 27.2 & 50.2 & 21.4 & 49.1 & 53.4 & 26.0 & 32.9 \\
wide-k32 & 11.91 & 76.0 & 41.6 & 73.3 & 31.0 & 61.7 & 68.2 & 47.8 & 50.3 \\
\bottomrule
\end{tabular}
\caption{Width configurations for both OLMoE checkpoints, with a training width of $k=8$. The suffix rn denotes additional renormalization of the selected expert weights to sum to one. HellaSwag is reported at $k=2$ and $k=4$ for the instruction-tuned checkpoint, and at $k=16$ and $k=32$ for the base checkpoint.}
\label{tab:f1}
\end{table*}

\end{document}